\documentclass[11pt]{article}

\usepackage[final]{acl}

\usepackage{times}
\usepackage{latexsym}
\usepackage{amsmath, amsfonts}
\usepackage{tikz}
\usetikzlibrary{positioning, arrows.meta}
\usepackage{booktabs}
\usepackage{tabularx}
\usepackage{enumitem}
\usepackage{orcidlink}

\usepackage[T1]{fontenc}
\usepackage[utf8]{inputenc}

\usepackage{microtype}

\usepackage{inconsolata}

\usepackage{graphicx}

\title{From Task Success to Productive Success: \\
Evaluating Human-AI Collaboration by Quality and Cost}

\author{
    Saki Imai \qquad Mert İnan \qquad Malihe Alikhani\\
    Northeastern University, Boston MA \\
    \texttt{\{imai.s, inan.m, m.alikhani\}@northeastern.edu}
}

\begin{document}
\maketitle
\begin{abstract}
AI productivity is often measured by task completion time, economic value, or improvements in outcome quality. However, these measures usually treat collaboration as a black box where they capture what output was produced, but not the interaction cost required to produce it. Motivated by economics literature, we introduce a productivity-oriented framework for evaluating human-AI collaboration as outcome quality relative to interaction cost. Across two datasets spanning four tasks, we show that: (1) sessions with identical quality ratings can differ by up to 70$\times$ in interaction cost; (2) quality-cost relationships vary by task, with some tasks rewarding extended interaction and others favoring fast convergence; (3) subjective user ratings are not reliable substitutes for productivity; and (4) productive sessions are characterized by agents probing earlier and users spending less effort repairing the interaction. By distinguishing productive success from costly success, our framework makes interactional cost visible and shows how dialogue analysis can inform the evaluation and design of AI systems. 
\end{abstract}

\section{Introduction}
\begin{figure}[t]
\centering
\resizebox{\columnwidth}{!}{
\begin{tikzpicture}[
    x=1cm, y=1cm,
    every node/.style={align=center},
    % Standard quadrant style
    quad/.style={
        draw=black!15,
        fill=black!3,
        line width=1pt,
        rounded corners=8pt,
        minimum width=4cm,
        minimum height=3.2cm,
        inner sep=8pt
    },
    % Highlight style for the focal point
    highlight/.style={
        draw=blue!50,
        fill=blue!5,
        line width=1.5pt
    }
]

% Axes lines
    \draw[->, >=Stealth, line width=1pt, black!60] (0,0) -- (9.6,0);
    \draw[->, >=Stealth, line width=1pt, black!60] (0,0) -- (0,8.6);

    % Axis titles (explicitly positioned to align perfectly with the centers of the quadrant gaps)
    \node[font=\Large\bfseries, text=black] at (4.7, -0.9) {Interaction Cost};
    \node[rotate=90, font=\Large\bfseries, text=black] at (-1.5, 4.4) {Output Quality};
    
    % Axis end labels (aligned with quadrant centers)
    \node[font=\normalsize\color{black!70}, anchor=north] at (2.5, -0.1) {Low};
    \node[font=\normalsize\color{black!70}, anchor=north] at (6.9, -0.1) {High};
    \node[font=\normalsize\color{black!70}, anchor=east] at (-0.1, 2.5) {Low};
    \node[font=\normalsize\color{black!70}, anchor=east] at (-0.1, 6.3) {High};

    % Quadrants
    \node[quad, highlight] (TL) at (2.5, 6.3) {
        \large\textbf{Productive Success}\\[4pt]
        \normalsize High quality, low cost
    };

    \node[quad] (TR) at (6.9, 6.3) {
        \large\textbf{Costly Success}\\[4pt]
        \normalsize High quality, high cost
    };

    \node[quad] (BL) at (2.5, 2.5) {
        \large\textbf{Cheap Failure}\\[4pt]
        \normalsize Low quality, low cost
    };

    \node[quad] (BR) at (6.9, 2.5) {
        \large\textbf{Unproductive Failure}\\[4pt]
        \normalsize Low quality, high cost
    };

    % Title
    % \node[font=\bfseries, anchor=south] at (4.7, 8.6)
    %     {Productivity distinguishes productive success from costly success};

\end{tikzpicture}
}
\vspace{-15pt}
\caption{Productivity distinguishes \emph{productive success} from \emph{costly success}. Task success captures output quality, but not the interaction cost required to achieve it. We define productive collaboration as high quality output achieved with relatively low interaction cost.}
\label{fig:productivity_quadrants}
\end{figure}
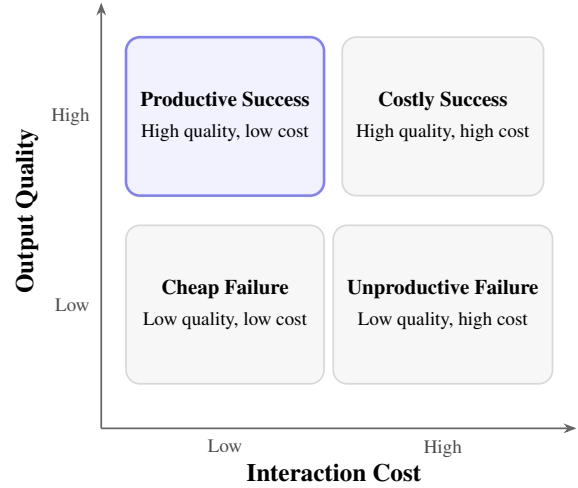

% concrete example
Two human-AI collaborations can achieve equally successful outcomes while requiring vastly different amounts of interaction from the user. In our data, for example, two travel planning sessions with the same quality rating differed by approximately 70 times in interaction cost. Task success treats these sessions as equivalent, but from the user's perspective they are not. Figure~\ref{fig:productivity_quadrants} illustrates this distinction: one is a \emph{productive success}, while the other is a \emph{costly success}.

% Consider two travel planning sessions, both rated as high quality. In one, the user and agent reached a satisfactory plan after a few hundred tokens of interaction. In the other, they required tens of thousands of tokens to reach an outcome with the same quality rating. From the perspective of task success, these sessions are \textit{indistinguishable}, but from the perspective of the user, they are \textit{significantly different}. Figure~\ref{fig:productivity_quadrants} illustrates this distinction: high quality outcomes can be either \emph{productive successes}, when they are reached with low interaction cost, or \emph{costly successes}, when they require high interaction cost.

As large language models become embedded in work practices, researchers have renewed efforts to measure AI productivity: what work AI systems can perform, how much economic value they may create, and whether they allow people to complete tasks faster or at higher quality. Recent work estimates the economic exposure of occupations to LLMs \cite{handa2025economic, tamkinmccrory2025productivity}, measures the time horizon over which agents can complete tasks autonomously \cite{measuring-ai-ability-to-complete-long-tasks, task-completion-time-horizons-of-frontier-ai-models, patwardhan2025gdpval}, and evaluates productivity gains in domains such as writing, coding, consulting, and customer support. These studies show that AI systems can affect productivity, but they often treat the interaction itself as a black box. They measure the final output, elapsed time, or economic value, without explaining \textit{how} the human and AI system reached that outcome.

% Recent interest in AI productivity has largely focused on the economic consequences of language models, by estimating monetary value \cite{handa2025economic, tamkinmccrory2025productivity}, labor substitution \cite{GPTs, webb2019impact, gans2026ring, handa2025economic}, or autonomous task completion \cite{measuring-ai-ability-to-complete-long-tasks, task-completion-time-horizons-of-frontier-ai-models, patwardhan2025gdpval}. However, these perspectives treat productivity as a property of the final output. They tell us relatively little about the \textit{interactional mechanisms} through which humans and AI systems become productive together, such as how users specify goals, how systems clarify, how misunderstandings are repaired, and how much effort is required to reach a satisfactory result.

% % explain the gap
% This is a limitation for evaluating LLM mediated collaboration. Unlike many productivity tools, LLMs collaborate through language. Users must specify goals, provide constraints, interpret responses, and repair misunderstandings. Agents can either reduce this burden by surfacing assumptions and asking useful clarifying questions, or increase it by forcing users to diagnose errors, restate goals, and manage overly long or misaligned responses. Two collaborations can therefore produce equally successful outputs while imposing very different \textbf{interactional costs} on the user.

We argue that productive human-AI collaboration should be evaluated as a relationship between outcome quality and interaction cost. This view builds on the principle of least collaborative effort \cite{clark1986referring}, which holds that collaborators coordinate not to minimize effort for one party alone, but to minimize the joint effort required to establish sufficient common ground \cite{clark1991grounding}. In human-AI settings, this distinction is especially important, where an agent can appear helpful while shifting substantial grounding work onto the user.

% Building on the principle of least collaborative effort \cite{clark1986referring} and theories of common ground \cite{clark1991grounding}, we view human-AI collaboration as a process of establishing sufficient mutual understanding for the task at hand. Productive collaboration depends not on whether grounding occurs, but on whether it helps the pair achieve a high quality outcome with less interactional burden.

To make this distinction measurable, we introduce a productivity oriented evaluation framework for human-AI collaboration that defines collaborative productivity as outcome quality relative to interaction cost. We validate the framework across two datasets, including a newly collected human-AI collaboration dataset with full interaction trajectories, submitted artifacts, and post task perception measures, for a visualization task.\footnote{Code and data to reproduce our work can be found at \url{https://github.com/sakimai/human-AI-productivity}}

% findings summary
Our analysis yields three main findings. First, sessions with identical quality ratings can differ by one to two orders of magnitude in interaction cost, variation that subjective user ratings do not consistently capture. Second, the relationship between cost and quality is task dependent: in related work writing, additional interaction tends to accompany better outcomes, while in visualization, higher quality sessions tend to converge with less interaction. These patterns suggest that productivity must be interpreted within task rather than treated as a universal preference for shorter conversations. Third, productive collaboration is not simply shorter interaction. Productive sessions are characterized by agents probing and clarifying earlier, while less productive sessions require users to spend more effort clarifying, redirecting, and repairing. This suggests that interaction can be productive when agent initiated grounding reduces later clarification, and correction, but costly when the burden of resolving ambiguity falls primarily to the user.

Our contributions are as follows:
\begin{enumerate}[noitemsep, nolistsep]
    \item We introduce a productivity oriented evaluation framework for human-AI collaboration as outcome quality relative to interaction cost (\S~\ref{ssec:metric}), and show that it captures variation not reflected in user subjective ratings (\S~\ref{ssec:divergence}).
    \item We show that collaborations with similar outcome can require dramatically different amounts of user effort (\S~\ref{ssec:variation}) and that tasks differ in whether additional interaction helps or hurts quality (\S~\ref{ssec:distributions}).
    \item We identify dialogue patterns associated with productive collaboration, showing that productive agents reduce user side grounding burden by front loading clarification (\S~\ref{ssec:dialogue}).
\end{enumerate} 

\section{Related Work}
\paragraph{Evaluating productivity of human-AI collaboration.}
% AI productivity studies measure outcome and not the interaction
A growing literature measures whether LLMs make people more productive, by measuring outcomes rather than interactions. Economic analyses estimate aggregate value and labor market effects \citep{handa2025economic, tamkinmccrory2025productivity, GPTs, webb2019impact, gans2026ring}, and benchmarks measure autonomous task completion \citep{measuring-ai-ability-to-complete-long-tasks, task-completion-time-horizons-of-frontier-ai-models, patwardhan2025gdpval, vidgen2025ai}. Empirical experiments with users report substantial speedups for professional writing \citep{noy2023experimental}, customer support \citep{brynjolfsson2025generative}, code generation \citep{peng2023impact, imai2022github, qian2024take}, ad creation \citep{ju2025collaborating}, and consulting \cite{dell2023navigating} with the largest gains among less experienced workers. Other studies report null or negative effects, where experienced developers were 19\% slower with AI while believing they were 20\% faster \citep{becker2025measuring}, and consultants who used AI on tasks outside its capabilities performed worse than those without it \citep{dell2023navigating}. These studies measure \textit{whether} AI helps on average, not \textit{how} language use during the interaction shapes the outcome. We close this gap by measuring collaborative productivity at the session level and analyzing the dialogue that produced each outcome. Earlier work combined task success and dialogue cost into a single score for spoken dialogue systems \citep{walker-etal-1997-paradise}. We adapt this framing to human-AI collaboration, but shift the question from comparing systems to explaining variation between sessions on the same task.

\paragraph{Evaluating human-AI collaboration.}
% human-AI evaluation is increasingly using interaction data but often ignoring the cost
Recent work has argued that evaluation should move beyond static, AI alone benchmarks \cite{lee2023evaluating, fragiadakis2024evaluating, xuan2026interactiveevaluationrequiresdesign}. Interaction centered datasets \cite{lee2022coauthor, chiang2024chatbot, zheng2024lmsys, zhao2024wildchat} and benchmarks \cite{ICLR2025_771155ab, chang-etal-2025-chatbench, li2024crowdsourced, NEURIPS2023_91f18a12} are aimed to evaluate model behavior in use. A related line of work evaluates \emph{human-agent collaboration}, where the AI system can communicate with the user while also taking actions in a shared task environment \citep{yao2025taubench, barres2025tau, shao2024collaborative, shen2025completion}. Recent work on LLM training further incorporates efficiency into multiturn objectives, penalizing excessive user reading and writing costs alongside task success and interaction quality \citep{wu2025collabllm}. Most existing evaluations still center on end-to-end task success rates \cite{shridhar2020alfworld, zhou2024webarena, xie2024osworld, jimenez2024swe}, and the cost of reaching that outcome is often ignored. By introducing a framework that jointly measures outcome quality and interaction cost, it allows us to distinguish productive and unproductive collaboration.

\paragraph{Grounding acts and positive friction.}
\label{ssec:commonground}
% grounding and positive friction explain why some cost is useful and some cost is wasteful
Collaboration requires participants to build a shared conceptual workspace \cite{roschelle1995construction}, and both speakers share the effort of building it. The \textbf{principle of least collaborative effort} \citep{clark1986referring} holds that participants minimize \textit{joint} rather than individual effort, and \citet{clark1991grounding} decompose this into formulation, reception, and repair costs distributed across both speakers. This framing motivates our framework: a productivity measure for human-AI collaboration must account for the burden agent output places on users. Recent work shows that LLMs do not naturally reproduce human grounding behavior. \citet{shaikh-etal-2024-grounding} find that LLM generations contain fewer grounding acts than human responses, often presuming common ground rather than actively constructing it. \citet{shaikh2025navigating} further show that grounding failures in human-LLM interaction can produce downstream rifts. In contrast, positive friction argues that some slowdowns improve reliability \cite{inan2025better}. Our framework reconciles these views, that positive friction is productive when it prevents larger downstream repair costs, and unproductive when it shifts repair burden to the user.
\section{Methods}
We evaluate human-AI collaboration at the session level. Each session produces a final output and an interaction trajectory. Our framework asks whether the session achieved high outcome quality relative to the interaction cost required to produce it. We first define the productivity framework, then describe how we instantiate quality, cost, and dialogue features in our two datasets.

% \subsection{Task Setup}
% Following prior work on human-AI collaboration \cite{shao2024collaborative, shen2025completion}, we model each collaborative task as a partially observable sequential decision process (POMDP) \cite{kaelbling1998planning} and study the interaction trace between a human user and an LLM. A task environment is
% \[
% \mathcal{E} = (\mathcal{S}, \mathcal{A}, \mathcal{T}, \mathcal{O}, \mathcal{U}),
% \]
% where \(\mathcal{S}\) is the task state, \(\mathcal{A}\) is the action space, \(\mathcal{T}\) is the transition function, \(\mathcal{U}\) is the instruction space, and \(\mathcal{O}\) is the observation space. Following Collaborative Gym, the effective instruction may evolve during collaboration: in addition to the initial request, the instruction space can include constraints, clarifications, and preferences that emerge through interaction. We study a logged collaboration trajectory
% \[
% \tau = \big[(\ell_1, a_1, o_1), \ldots, (\ell_T, a_T, o_T)\big],
% \]
% where \(\ell_t \in \{H, A\}\) denotes whether the acting party at step \(t\) is the human or the agent, \(a_t \in \mathcal{A}\) is the action, and \(o_t \in \mathcal{O}\) is the resulting observation. Each trajectory yields a final artifact \(y(\tau)\). We distinguish the \textbf{quality} ($Q(y(\tau)$) of the final artifact from the \textbf{cost} ($C(\tau)$) of the interaction required to obtain it. Our goal is to understand why trajectories with similar final outcomes can differ substantially in the cost they impose on the user.

\subsection{Collaborative Productivity Framework}
\label{ssec:metric}

We define collaborative productivity as outcome quality relative to interaction cost. This follows the productivity intuition of output relative to input \citep{solow1957technical, schreyer2001measuring} in economics, but adapts it to human-AI collaboration,\footnote{We use \textit{human-AI collaboration} as an umbrella term covering both chat-only and tool-using LLM agents.} where the relevant input is the interactional effort required to reach an outcome.

This formulation is motivated by theories of grounding in communication. Collaborative work requires participants to establish sufficient common ground for the task at hand. \citet{clark1991grounding} distinguish several costs involved in this process, including formulation costs for producing utterances, reception costs for interpreting them, and repair costs for resolving misunderstandings. In human-AI collaboration, these costs are distributed across the user and the agent, where users write prompts, read and interpret model outputs, and repair when the interaction goes off track. We therefore operationalize interaction cost as a property of the full trajectory rather than only the user's messages. Our framework also builds on PARADISE, which evaluates spoken dialogue agents by combining task success with dialogue costs \citep{walker-etal-1997-paradise}. PARADISE showed that dialogue evaluation should separate what was accomplished from the costs incurred in accomplishing it, and that these quantities can be combined into a performance function. We adapt this idea from spoken dialogue systems to human-AI collaboration.

Let $\mathcal{Y}$ denote the space of artifacts a collaborative session can produce and $\mathcal{T}$ the space of interaction trajectories. We define a \textit{quality function} $Q : \mathcal{Y} \to \mathbb{R}$ scoring the artifact and a \textit{cost function} $C : \mathcal{T} \to \mathbb{R}_{\geq 0}$ scoring the interaction. The collaborative productivity of session $i$ on a task is
\begingroup
\setlength{\abovedisplayskip}{4pt}
\setlength{\belowdisplayskip}{4pt}
\setlength{\abovedisplayshortskip}{4pt}
\setlength{\belowdisplayshortskip}{4pt}
\begin{equation*}
P_z(i) = z(Q_i) - z(C_i)
\end{equation*}
\endgroup
where $z(\cdot)$ denotes within task z-score normalization and $Q_i, C_i$ are the quality and cost of session $i$. We standardize $Q$ and $C$ because tasks differ in both quality scales and expected interaction length, so raw scores are not directly comparable across tasks. Following multi-attribute utility approaches \citep{keeney1993decisions}, we combine the standardized attributes additively, treating quality as a benefit and interaction cost as a penalty. Thus, $P_z$ can be interpreted as an index that shows whether a session achieved higher quality for that task while requiring lower interaction cost. In Appendix~\ref{ssec:validate}, we report
sensitivity analyses varying the relative weight on quality and cost and show that the main qualitative findings are stable across weights.

% We test robustness to a $\lambda$ weighted family $P^{\lambda}_i = z(Q_i) - \lambda \, z(C_i)$ for $\lambda \in [0.5, 2.0]$ in \S\ref{ssec:validate}.

% \[
% P(\tau) = f\big(Q(y(\tau)), C(\tau)\big),
% \]
% where \(f\) increases with quality and decreases with cost. Combining quality and cost into a single score requires a principled way to balance two incommensurable quantities. Multi-attribute utility theory \citep{keeney1993decisions} addresses this exact problem, that when a decision involves attributes measured on different scales, each attribute is normalized to a common scale before combination, and weighted combinations of normalized attributes yield comparable scores across alternatives.
% We instantiate this approach with the simplest reasonable choice, where dimensions are standardized within each task and combined with equal weight
% \[
% P_i = z(Q_i) - z(C_i)
% \]
% where $z(\cdot)$ is a Z score normalization function computed within each task, so that productivity measures how productive a session compares to other sessions on the same task. Equal weighting reflects treating quality and cost as co-equal dimensions of productivity, in line with the joint effort principle in \S~\ref{ssec:commonground}. We test robustness to a family of $\lambda$ weighted formulations $P^{\lambda} = z(Q) - \lambda z(C)$ for $\lambda \in [0.5, 2.0]$. Stability of rank orderings across this range is evidence that the metric captures the quality-cost relationship than an artifact of the specific $\lambda = 1.0$ choice (\S~\ref{ssec:validate}).

\paragraph{Quality as a task dependent design choice.}
The appropriate quality function ($Q$) depends on the task and evaluation goal. In code generation, the standard target is functional correctness, usually reported as pass@k \cite{chen2021evaluating} or test case pass rate \cite{hendrycks2021measuring}. For high stakes domains, such as medical \cite{ayers2023comparing, tu2025towards} or mental health \cite{scholich2025comparison, foyen2025artificial} support, expert ratings may be appropriate because correctness and harm require domain judgment. For service oriented tasks, user ratings may be appropriate when the target outcome is whether the user's goal was satisfied \cite{budzianowski2018multiwoz}. For open-ended but rubric evaluable tasks, LLM-as-a-judge rubric based scoring provides a consistent way to assess artifact quality across sessions.

\paragraph{Interaction cost as grounding burden.}
Cost measures should reflect the interactional burden imposed by the collaboration. User tokens approximate formulation effort, agent tokens approximate reception burden, and turns or elapsed time capture coarser interaction overhead. Repair costs are not added as a separate term in our primary measure because they accumulate through the trajectory itself where repeated clarification and reformulation increase the number of tokens. In this sense, token and turn based costs operationalize the idea from grounding theory that formulation, reception, and repair effort accumulate over the interaction.

\subsection{Data}
\label{ssec:data}
We instantiate the framework on two datasets spanning four tasks: three structured human-agent tasks from Collaborative Gym and one newly collected human-LLM visualization task.

\paragraph{CoGym}
We use human-agent collaboration sessions from the CoGym framework \cite{shao2024collaborative} spanning three diverse tasks: travel planning (n = 112), tabular analysis (n = 69), and related work synthesis (n = 47). In these tasks, users collaborate with an LLM agent that can act in a shared environment, such as searching databases, editing a shared document, or executing code. Sessions were collected from real users interacting with LLM agents (GPT-4o, Gemini 2.0 Flash) through a web interface \cite{hurst2024gpt, team2023gemini}. We restrict primary analyses to sessions producing a final artifact (n=184), and provide additional details of each task in Appendix~\ref{app:cogym}.

\paragraph{Visualization} We additionally collect a new human-LLM collaboration dataset of 42 sessions in which participants used GPT-5.1 \cite{singh2025openai} to analyze the WildChat-1M dataset \cite{zhao2024wildchat}. Participants were asked to formulate a hypothesis about the dataset, test it using Python, and submit a visualization with supporting code and a short interpretation. The task is open-ended but the final output can still be evaluated with a rubric for relevance, correctness, and evidential support. We log the full chat trajectory, submitted code, final visualization, and post task survey responses. After submission, participants rated their confidence in the output, the usefulness of the AI, and perceived interaction productivity. This dataset complements CoGym by covering a human-AI collaboration task without environment actions. We provide additional details on the study setting, custom platform, and procedure in Appendix~\ref{app:visualization}.

\paragraph{Defining Q and C.} Across both datasets, we use LLM-as-judge rubric scored task performance as the primary quality measure $Q$ \citep{chiang2024chatbot}, using the task specific rubrics reported in Appendix~\ref{app:rubrics}. This choice reflects our goal of measuring productivity as the quality of the final artifact relative to the interaction cost required to produce it. On the other hand, user Likert ratings capture how participants perceived the outcome or interaction. We therefore reserve them for comparison analyses rather than using them to define $Q$. This separation allows us to ask whether perceived productivity aligns with an independently scored quality-cost tradeoff, and avoids circularity in later analyses. Moreover, this avoids using the same user survey both to define $Q$ and to test whether users perceived the interaction as productive \citep{podsakoff2003common}. In our data, subjective Likert ratings are strongly correlated within each dataset, so we keep them as comparison measures rather than the primary quality measure (Appendix~\ref{app:likert_corrs}). Our primary cost measure is weighted token cost:
\[
C = \text{user tokens} + 0.5 \cdot \text{agent tokens}
\]
User tokens approximate formulation effort, while agent tokens approximate reception burden \citep{clark1991grounding}. We assign agent tokens a lower weight because reading is typically faster than composing text \cite{brysbaert2019many, 10.1145/302979.303160}, while still treating long model outputs as a cost imposed on the user. Because no token based measure can fully capture effort such as cognitive load and attention, we test whether our conclusions depend on this particular operationalization. Appendix~\ref{ssec:validate} repeats the analyses with user tokens only, total tokens, turn count, and alternative agent token weights; the main conclusions remain qualitatively stable.

\paragraph{Human validation.}
To validate the LLM-as-judge quality scores, two authors independently scored a subset of sessions using the same task specific rubrics. For CoGym, we sample 20 sessions from each task and across the range of LLM-rubric scores; for the visualization dataset, we score all 42 sessions. Human scores correlate strongly with LLM-as-judge scores (CoGym $\rho=.68$; visualization $\rho=.74$), with high human--human agreement ($\rho=.81$). This suggests that the primary quality signal used in $P_z$ tracks human rubric judgments.

\subsection{Grounding and Friction Features}
\label{ssec:features}

To characterize productive collaboration, we annotate each utterance using two taxonomies: grounding acts \citep{shaikh2025navigating} and positive friction movements \citep{inan2025better}. Grounding acts characterize whether and how participants establish mutual understanding; friction movements characterize deliberate slowdowns that surface assumptions or invite reflection. The two taxonomies capture different aspects of collaborative discourse.

\paragraph{Grounding acts.} 
We adapt the taxonomy of \citet{shaikh2025navigating}, which distinguishes acts that \textit{advance} grounding (\textit{next turn}, \textit{acknowledge}, \textit{follow-up}), acts that \textit{signal ambiguity} (\textit{clarification}, \textit{overresponse}), and acts that \textit{address} grounding failures (\textit{repair}, \textit{reformulation}, \textit{restart}). User and agent acts are labeled separately, since the same surface act carries different functional weight depending on the speaker (e.g., an agent acknowledgment vs. a user acknowledgment). An utterance may receive multiple labels.

\paragraph{Positive friction movements.}
We adapt the five category positive friction taxonomy from \citet{inan2025better}: \textit{assumption reveal} (surfacing beliefs about the environment, the interlocutor, or the task), \textit{reflective pause} (verbal or behavioral signals of internal deliberation), \textit{reinforcement} (restating a prior utterance for emphasis), \textit{overspecification} (providing more information than requested), and \textit{probing} (questioning to redirect or clarify). Each utterance receives a single label from these five categories or \textit{Not Friction} if none applies. We retain the single label scheme of the original taxonomy.

\paragraph{Annotation.}
We use GPT-5.1 at temperature 0 to label each utterance with grounding acts and positive friction movements, using the annotation guidelines from \citet{shaikh2025navigating} and \citet{inan2025better}. We validate these labels on a subset of 20 sessions, balanced across dataset and productivity quartile. Two human annotators independently label the same utterances. Agreement was substantial for positive friction labels (human--human $\kappa=.72$; model--human $\kappa=.64$) and acceptable for grounding acts (human--human micro-F1 $=.76$; model--human micro-F1 $=.68$). 
% We use the model generated labels for the full-corpus analysis and provide label-wise validation results in Appendix~\ref{app:human_validation}.
\section{Findings}

\begin{figure*}
    \centering
    \includegraphics[width=0.9\linewidth]{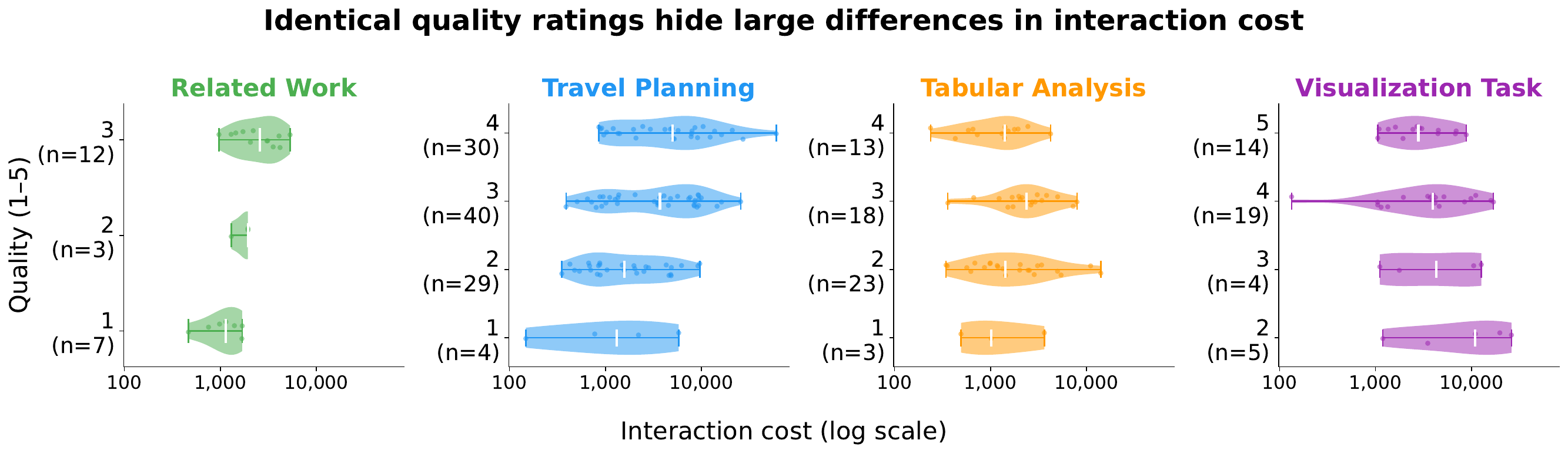}
    \caption{
          Identical quality ratings can hide large differences in interaction cost. Each panel shows the distribution of interaction cost within quality buckets for one task. Cost is shown on a log scale. Wide horizontal spread within a quality bucket indicates that task success alone cannot distinguish \emph{productive success} from \emph{costly success}.
    }
    \vspace{-5pt}
    \label{fig:cost_within_quality}
\end{figure*}

\begin{figure*}
    \centering
    \includegraphics[width=0.9\linewidth]{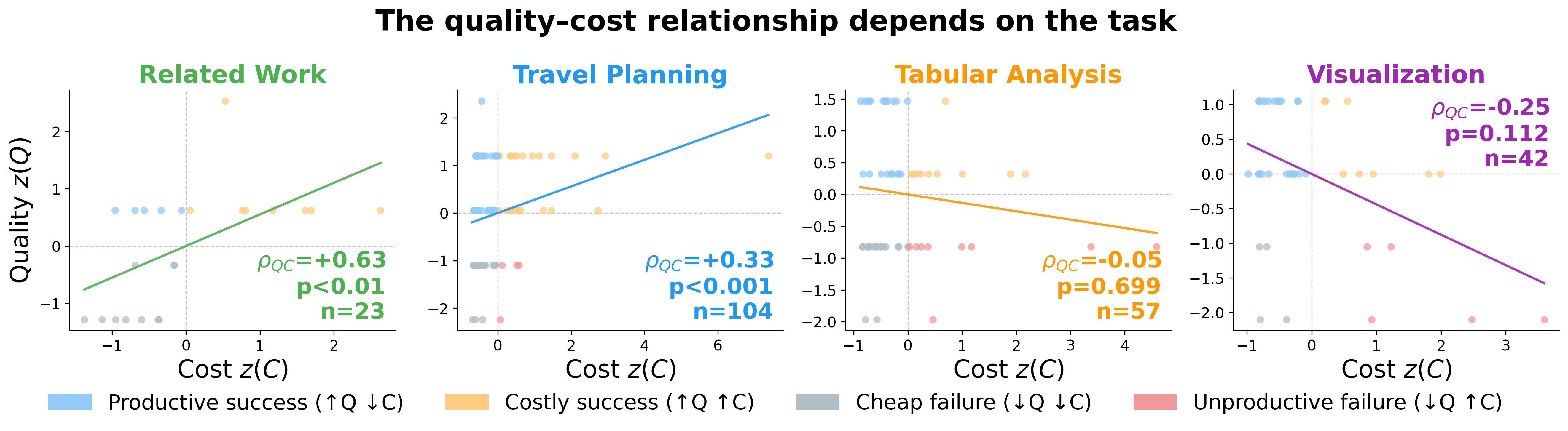}
    \caption{
        Task specific relationships between outcome quality and interaction cost. Each point is a session, with quality and cost standardized within task; dashed lines indicate task means. The pattern shows that interaction cost is not uniformly beneficial or harmful, where additional interaction accompanies higher quality in some tasks. 
    }
    \label{fig:quality_cost_by_task}
\end{figure*}

We organize the results around four questions. First, does task success hide interaction costs? (\S~\ref{ssec:variation}) Second, does the quality-cost relationship vary by task? (\S~\ref{ssec:distributions}) Third, do subjective ratings capture the same quality-cost tradeoff? (\S~\ref{ssec:divergence}) Finally, what dialogue patterns characterize productive collaboration? (\S~\ref{ssec:dialogue}) These analyses show that productivity is not simply shorter interaction. It depends on whether interaction cost helps produce quality, and on who bears the grounding work required to reach that quality.

\subsection{Cost Variation Within Quality Levels}
\label{ssec:variation}
% i'd like this section to motivate the need for our metric
Identical quality ratings can require very different amounts of interaction.  Figure~\ref{fig:cost_within_quality} plots interaction cost within each quality bucket separately for all three CoGym tasks and the visualization task. Across tasks, sessions with the same rating often differ by one to two orders of magnitude in cost.

The clearest example appears in travel planning. Among completed sessions in the top quality bucket, interaction costs range from 854 to 60,324 tokens, a 70.6 times difference. Similar within quality variation appears in tabular analysis, related work, and visualization. This motivates measuring productivity as quality relative to cost since task success alone cannot distinguish \emph{productive success}, where high quality is reached with low interaction cost, from \emph{costly success}, where similar quality requires substantially more interaction. The same pattern holds when quality is measured with human Likert outcome rating rather than LLM rubric scores (Appendix~\ref{app:likert_variation}).

\subsection{Task Structures Shape Quality \& Cost}
\label{ssec:distributions}

Interaction cost is not uniformly helpful or harmful; its relationship to quality depends on the task. We compute the within task Spearman correlation between standardized quality $z(Q)$ and standardized cost $z(C)$, and Figure~\ref{fig:quality_cost_by_task} visualizes the relationship at the session level.

Related work shows the strongest positive association ($\rho_{QC}=+0.55$), suggesting that additional interaction often supports more complete synthesis. Travel planning is also positive but weaker ($\rho_{QC}=+0.27$), consistent with a task where iteration helps but gains diminish as plans converge. In contrast, tabular analysis shows little evidence of a positive quality-cost relationship ($\rho_{QC}=-0.13$) where additional exchanges do not reliably improve quality. This interpretation is consistent with works showing that data analysis is iterative and often repetitive \cite{wongsuphasawat2019goals, kandel2012enterprise}. Visualization shows the most negative association ($\rho_{QC}=-0.48$), suggesting that high quality sessions depend less on iteration and tend to converge quickly on a hypothesis and visualization.

These differences matter for evaluation. A universal penalty on interaction length would mischaracterize tasks like related work, where more interaction can reflect useful elaboration. Conversely, rewarding longer interaction would mischaracterize tasks where additional exchanges reflect debugging, uncertainty, or delayed convergence. Productivity therefore needs to be interpreted within a task rather than a universal preference for brevity.

% \begin{table}[t]
% \centering
% \small
% \setlength{\tabcolsep}{4pt}
% \begin{tabularx}{\columnwidth}{l r X}
% \toprule
% \textbf{Task} & \textbf{$\rho_{QC}$} & \textbf{Takeaway} \\
% \midrule
% Related work    & $+0.55$ & More interaction often improves quality. \\
% Travel planning & $+0.27$ & Iteration helps, but gains diminish. \\
% Tabular analysis & $-0.13$ & Extra exchanges do not reliably improve quality. \\
% Visualization   & $-0.48$ & High-quality sessions tend to converge quickly. \\
% \bottomrule
% \end{tabularx}
% \caption{Task structure shapes the relationship between interaction cost and outcome quality. $\rho_{QC}$ is the within task Spearman correlation between standardized quality $z(Q)$ and standardized interaction cost $z(C)$.}
% \label{tab:pz_descriptives}
% \end{table}

% \begin{figure}[t]
% \centering
% \includegraphics[width=\columnwidth]{figures/fig_pz_taskz_kde.png}
% \caption{Kernel density estimates of $P_z$ by task. Distributions retain task specific shape after per task z-scoring; the visualization task is right shifted with a heavy left tail.}
% \label{fig:pz_kde}
% \end{figure}

\subsection{Subjective Ratings Do Not Reliably Capture Productivity}
\label{ssec:divergence}

If subjective ratings captured productivity, then users should penalize costly sessions when quality is held fixed. We test this by computing, for each subjective measure $S$, the partial Spearman correlation $\rho(C,S\mid Q)$ between interaction cost $C$ and the subjective rating, controlling for quality $Q$. A productivity sensitive subjective measure should show a negative partial correlation, where among sessions of similar quality, higher cost should correspond to lower ratings. Across six subjective measures in two datasets, this pattern appears reliably in only one case (Table~\ref{tab:subjective_dissociation}).

CoGym \texttt{satisfaction} shows the predicted cost penalty, both unconditionally ($\rho=-0.22$, $p<0.01$) and after controlling for quality ($\rho=-0.24$, $p<0.01$). In contrast, CoGym \texttt{outcome quality} and \texttt{communication rating} are cost blind: holding quality fixed, longer or more costly interactions are not rated lower. But, the visualization study shows \texttt{Speed} is weakly cost-blind, \texttt{effectiveness} is cost-rewarding, and \texttt{confidence} is positively associated with cost even after controlling for quality ($\rho=+0.41$, $p<0.01$).

These divergences suggest that subjective prompts elicit different constructs. Quality focused prompts (\texttt{outcome quality}, \texttt{communication rating}) anchor users to the final artifact, where interaction cost may be less salient. Experience focused prompts can split into two interpretations, where satisfaction appears to treat cost as negative experience, while confidence may treat cost as investment or commitment. Subjective ratings are therefore useful, but they are not interchangeable proxies for productivity.

\begin{table}[t]
\centering
\resizebox{\columnwidth}{!}{
\setlength{\tabcolsep}{4pt}
\begin{tabular}{llrrr}
\toprule
\textbf{Dataset} & \textbf{Measure} & $\rho(Q, S)$ & $\rho(C, S \mid Q)$ & $n$ \\
\midrule
CoGym    & outcome quality      & $+0.14$       & $-0.04$       & 155 \\
CoGym    & comm. rating & $-0.09$       & $-0.14$       &  41 \\
CoGym    & satisfaction          & $+0.03$       & $-0.24^{**}$  & 184 \\
Visualization & speed                 & $-0.02$       & $-0.07$       &  42 \\
Visualization & effectiveness         & $-0.05$       & $+0.13$       &  42 \\
Visualization & confidence            & $-0.00$       & $+0.41^{**}$  &  42 \\
\bottomrule
\end{tabular}
}
\caption{Partial Spearman correlations of interaction cost $C$ with subjective measures $S$, controlling for LLM rubric quality $Q$. CoGym measures are either cost-blind or cost-penalizing; Visualization metrics are either cost-blind or cost-rewarding. *$p<0.05$, **$p<0.01$.}
\vspace{-10pt}
\label{tab:subjective_dissociation}
\end{table}

\subsection{Productive Collaboration Shifts Interactional Labor to the Agent}
\label{ssec:dialogue}

\begin{figure*}[t]
    \centering
    \includegraphics[width=0.8\textwidth]{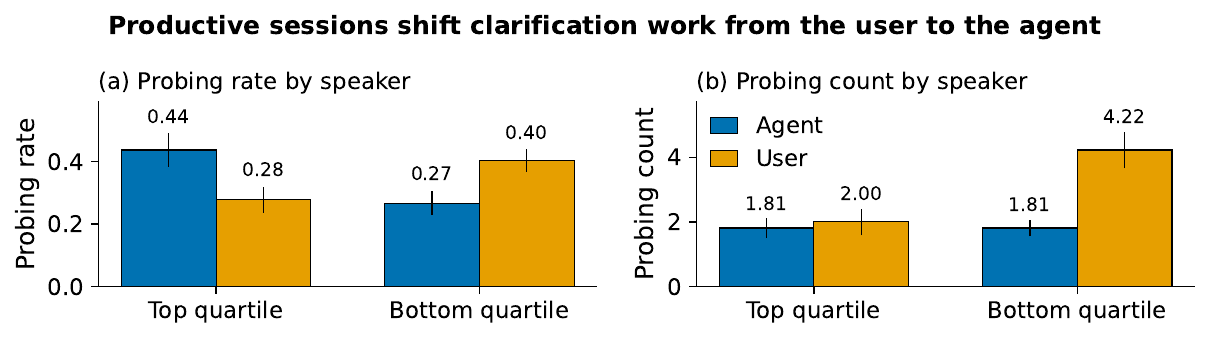}
    \caption{Agent and user probing in top vs.\ bottom $P_z$ quartile (n=59 each). 
    (a)~Rate per turn: productive sessions show a labor crossover, with the agent probing more and the user probing less. 
    (b)~Counts per session: agent probe \textit{counts} are identical, but user probe counts more than double in unproductive sessions. Error bars: $\pm$1 SEM.}
    \vspace{-5pt}
    \label{fig:probing}
\end{figure*}

We next ask what productive collaboration looks like in the dialogue itself. We compare the top and bottom $P_z$ quartiles (n=59 each) on grounding acts and positive friction movements. For each feature, we report Cohen's $d$ for the difference in means between quartiles, where positive values indicate features that are more frequent in productive sessions. Significance is assessed with Mann Whitney $U$ tests; full feature results are in Appendix~\ref{app:dialogue_full}.

\paragraph{Friction is not inherently unproductive; its speaker matters.}
The rate of any friction per turn is nearly identical between productive and unproductive sessions (0.77 vs.\ 0.79, $d=-0.31$, n.s.). However, unproductive sessions contain more total friction movements because they are longer (8.0 vs.\ 14.1, $d=-0.65$, $p<0.001$). The difference between these sessions is therefore not how often friction occurs, but \textit{who} produces it. 

The clearest example is probing (Figure~\ref{fig:probing}). Productive sessions devote 44\% of agent turns to probing, compared to 27\% in unproductive sessions ($d=+0.73$, $p<0.001$). Across the full sample, agent probing rate is positively correlated with $P_z$ ($\rho=+0.19$, $p<0.01$, n=233). This effect is concentrated in tasks where iterative clarification is useful, such as related work ($\rho=+0.35$) and travel planning ($\rho=+0.24$), consistent with the task level patterns in \S\ref{ssec:distributions} (Table~\ref{tab:per_task_probing}). Agent probe \textit{counts} are identical across quartiles ($\bar{n}=1.81$ in both; Figure~\ref{fig:probing}b), which shows that productive sessions reach the same amount of clarification in roughly half the number of turns. The pattern is stronger early in the session where 48\% of agent turns in the first 30\% of productive sessions contain probing, vs.\ 31\% in unproductive ones. User probing shows the opposite pattern ($d=-0.52$, $p<0.01$), suggesting that when the agent does not probe, users must do the clarification work themselves. Assumption reveal follows the same trend, concentrated on the user side in unproductive sessions ($n=0.31$ vs.\ $1.37$, $d=-0.57$).

\paragraph{Grounding acts show the same asymmetry.}
User repair (0.22 vs.\ 0.83, $d=-0.58$), clarification (0.27 vs.\ 0.63, $d=-0.49$), follow-up (3.05 vs.\ 6.02, $d=-0.55$) are all elevated in unproductive sessions. By contrast, user acknowledgment and agreement rates run roughly 3$\times$ higher in productive sessions ($d=+0.58$ for both, $p<0.05$). Productive collaboration therefore looks like \textit{agent probes $\rightarrow$ user confirms}; unproductive collaboration leaves the grounding work to the user to repair, clarify, and probe. This matches the principle of least collaborative effort \citep{clark1986referring,clark1991grounding}. Agent side friction can be productive when it surfaces ambiguity early and prevents downstream repair. The same friction becomes costly when the user must perform it later as repair work.

\section{Discussion}
Our findings suggest that productive human-AI collaboration is about how interactional work is distributed across the collaboration.

\paragraph{Productive collaboration depends on who bears the grounding work.}
Our dialogue analysis shows that productive collaboration is not simply less interactive or less frictional. Productive and unproductive sessions contain similar rates of friction; what differs is who performs the grounding work (\S\ref{ssec:dialogue}). In productive sessions, agents probe earlier and users confirm; in unproductive sessions, users repair, clarify, and probe the agent. Calls for seamless interaction often imply that friction should be minimized, while positive friction argues that strategic slowdowns can improve reliability \citep{inan2025better}. Our results suggest that friction is productive when the agent pays a small clarification cost up front, and costly when unresolved ambiguity later becomes user-side repair. \textit{This implies that agents should be designed to front-load clarification, surface assumptions, and ask targeted probes before acting, rather than waiting for users to initiate repair.}

\paragraph{Productivity is task relative.}
The quality-cost relationship varies substantially across tasks, from positive in related work synthesis ($\rho_{QC}=+0.55$) to negative in visualization ($\rho_{QC}=-0.48$; \S\ref{ssec:distributions}). This means there is no universal effort to outcome curve for human-AI collaboration. In some tasks, additional interaction reflects useful elaboration, exploration, or synthesis; in others, it may reflect delayed convergence, debugging, or confusion. Evaluation protocols that uniformly penalize interaction length may therefore underrate systems on tasks where extended interaction is productive, while protocols that reward engagement may overrate systems on tasks where high quality outcomes should converge quickly. \textit{This implies that productivity should be evaluated and interpreted within a task before comparing across tasks or systems.}

\paragraph{Subjective ratings measure different constructs.}
Subjective ratings are valuable, but they are not interchangeable measures of productivity. In our data, most subjective measures do not penalize interaction cost once quality is held fixed, and some move in the opposite direction (\S\ref{ssec:divergence}). This suggests that quality focused prompts direct users to the artifact, satisfaction may integrate cost as negative experience, and confidence may reflect commitment. As a result, two studies could reach different conclusions about the same system depending on whether they ask users about satisfaction, confidence, usefulness, or speed. \textit{This has implications in studies of human-AI collaboration which should specify what each subjective prompt is intended to measure, and should avoid treating perceived quality, usefulness, confidence, and productivity as interchangeable outcomes.}

\textbf{Productivity is conditional on the evaluation objective.}
Our framework does not assume that interaction is inherently costly or that shorter interactions are always preferable. Interaction cost is meaningful only relative to the outcome the collaboration is intended to achieve, and the quality function (Q) must therefore reflect the values relevant to that task. For example, in an educational setting, immediately giving a student the correct answer may minimize interaction while undermining the objective of learning. If learning is the intended outcome, $Q$ should instead capture outcomes such as retention, transfer, or independent problem solving. We therefore view collaborative productivity as efficiency conditional on task relevant goals and not as a  prescription to minimize interaction.

\section{Conclusion}
Task success tells us whether a human-AI collaboration reached a desired outcome, but not what it cost the user to get there. We introduced a productivity oriented framework for evaluating collaboration as outcome quality relative to interaction cost, and applied it across four tasks from two datasets. Our results show that identical quality ratings can hide large differences in interaction cost, that the quality-cost relationship varies by task, and that subjective ratings do not consistently capture this tradeoff. Dialogue analysis further shows that productive collaboration shifts grounding work toward the agent. Productive collaboration is not just successful collaboration made shorter. It is successful collaboration in which interactional effort is spent where it contributes meaningfully to the outcome rather than being unnecessarily transferred to the user. These findings motivate the design of agents that surface ambiguity and probe when appropriate to reduce unnecessary user repair.
\section*{Limitations}
\paragraph{Interaction cost is an approximation of user effort.}
We use weighted token cost as our primary measure, treating user tokens as a proxy for formulation effort and agent tokens as a proxy for reception burden. This operationalization captures part of the interactional burden, but it does not directly measure cognitive load, attention, frustration, multitasking, or the effort required to evaluate whether an answer is correct. 

\paragraph{Dialogue patterns are correlational.}
Our dialogue analysis identifies features associated with productive sessions, such as agent probing and reduced user repair. However, these results are observational. Future experiments should test whether prompting or training agents to front load clarification improves productivity without introducing unnecessary friction.

\paragraph{Generalizability is limited by tasks.}
We evaluate four tasks across two datasets. These settings provide useful variation, but they do not cover the full range of human-AI collaboration. Productivity may behave differently in long horizon work, creative collaboration, or domains where trust, safety, and accountability matter more than speed.

\section*{Use of AI Assistants}
The authors used AI assistants for language polishing after the initial draft. The authors verified and edited all generated text and remain responsible for the content of the paper. 

\section*{Acknowledgments}
This research was supported in part by a grant from the Institute for Information, the Internet, and Democracy (IIID) at Northeastern University. We thank Asteria Kaeberlein for their helpful feedback.

\bibliography{custom}

\appendix
\section{CoGym Dataset Task Details}
\label{app:cogym}
We use three task environments from Collaborative Gym (CoGym): travel planning, related work writing, and tabular analysis. CoGym is designed for human-agent collaboration in shared task environments, where both the human and the agent can communicate and interact with task specific tools. 

\paragraph{Travel planning.}
In the travel planning task, the human and agent collaborate to produce a detailed itinerary that satisfies a user’s travel goals and constraints. The task is challenging because the initial request may underspecify important preferences, such as preferred activities, destinations, budget constraints, accommodation needs, or transportation choices. The agent can use travel-related search tools, including city, attraction, restaurant, flight, and accommodation search, and can update a shared itinerary editor. The final artifact is the completed travel plan. 

\paragraph{Related work writing.}
In the related work task, the human and agent collaborate to write a related work section for a given research topic. The agent can search for relevant papers, add or remove papers from a shared library, transfer selected papers into a draft, and update a shared text editor. The final artifact is the related work section. 

\paragraph{Tabular analysis.}
In the tabular analysis task, the human and agent collaborate to derive analytical insights from provided tabular data. The environment includes shared access to the data, a Jupyter notebook for code execution, and a text editor for documenting findings. The agent can execute notebook cells and update the written analysis. The final artifact is an analytical report or finding supported by computations over the table.

These three tasks provide variation in the role of interaction. Travel planning emphasizes latent preferences and constraint satisfaction; related work writing emphasizes synthesis, domain expertise, and iterative refinement; and tabular analysis emphasizes tool use, code execution, and evidence generation. This variation is important for our analysis because productivity should not be interpreted as a universal preference for shorter interactions. Instead, the relationship between quality and cost depends on the task structure.

\section{Visualization Dataset Details}
\label{app:visualization}
\paragraph{Task.}
The visualization dataset was collected from an IRB approved study. Participants completed a structured exploratory analysis task using WildChat-1M \citep{zhao2024wildchat}, a large scale dataset of real world ChatGPT interactions. The task required participants to formulate a testable hypothesis about the dataset, write or adapt Python code to analyze the data, produce a visualization, and submit a short written interpretation of their findings. This design was intended to capture a realistic human--LLM workflow in which users must move from problem formulation to analysis, visualization, and explanation, rather than simply rate model outputs or complete isolated microtasks.

\paragraph{Participants.}
Participants were university students. This population is appropriate for the task because the study focuses on how users collaborate with an LLM to complete an open-ended analytical task. The study therefore captures collaboration in a setting where participants had a genuine task objective and were expected to produce a usable final artifact. In total, we collected 42 completed sessions. 

\paragraph{Interface and model.}
Participants interacted with the LLM through a custom web platform designed for the study. The interface presented the task instructions, provided access to the chat based assistant, and collected the final submission. The assistant was powered by GPT-5.1. The platform logged the full interaction trajectory, including user messages and model responses, so that each final artifact could be analyzed together with the process that produced it. 
% A screenshot of the platform is shown in Figure~\ref{fig:visualization_platform}.

\paragraph{Collected data.}
For each session, we collect the full chat transcript, timestamps if available, submitted code, final visualization, written interpretation, rubric score, and post task survey responses.

\paragraph{Privacy and release.}
Because the dataset contains natural language interactions with an LLM, chat logs may include incidental personal information or sensitive content entered by participants. Before any release, we will remove direct identifiers, anonymize participant IDs, and filter or redact potentially identifying content.

\section{Task Quality Rubrics}
\label{app:rubrics}

We evaluate the final artifact from each collaboration session using task specific grading rubrics. Each rubric assigns an integer score from 1 to 5, where higher scores indicate higher task quality. The judge is given the original task request and the final artifact produced by the human-AI team. For tabular analysis, the judge is also given excerpts from the Jupyter execution history so that the final report can be evaluated against the computations used to produce it. The judge is instructed to provide a short justification and then conclude with a final numeric score.

For the related work writing task, we reuse the rubric provided by Collaborative Gym in Figure 13 of \citet{shao2024collaborative}. We use custom rubrics for travel planning, tabular analysis, and the visualization task, shown below.

\subsection{Travel Planning Rubric}
\label{app:rubric_travel}

The travel planning rubric evaluates whether the final itinerary satisfies the user's request, covers the requested dates, includes concrete travel entities, and is internally realistic.

\begin{quote}
\textbf{Scoring Rubric for Travel Planning}

The evaluator is given: (a) the user's travel-planning request, and (b) the final travel plan that the human-AI team produced, defined as the contents of the shared travel-plan editor at the end of the session.

\textbf{Evaluation Criteria}

\textbf{Score = 1.}
The plan is missing entirely, gibberish, or unrelated to the user's request, including the origin, destination, or dates.

\textbf{Score = 2.}
The plan exists but only sketches one or two days. Required slots, such as transportation, accommodation, attraction, breakfast, lunch, and dinner, are missing for most days. Specifics such as flight numbers, restaurant names, or hotel names are absent or appear fabricated.

\textbf{Score = 3.}
The plan covers all requested days but is partially incomplete. Some meals or attractions are left blank; flight numbers, accommodations, or restaurants are present but generic; or there are commonsense issues, such as destination mismatches, impossible timing, or missing intercity transportation.

\textbf{Score = 4.}
The plan covers every day and fills most meal, attraction, transportation, and accommodation slots with concrete entities, such as named flights, hotels, restaurants, or attractions. The itinerary is internally consistent, with transportation between cities scheduled before the day's activities. Only minor gaps remain.

\textbf{Score = 5.}
The plan is complete and realistic for every day. It includes named flights with flight numbers, named hotels, named restaurants for each meal, and named attractions. It respects the user's preferences, including budget, interests, and number of travelers; places transportation between cities correctly; and contains no obvious commonsense violations.

The evaluator provides 2--4 sentences of justification and concludes with: \textit{Therefore, the final score is N.}
\end{quote}

\subsection{Tabular Analysis Rubric}
\label{app:rubric_tabular}

The tabular analysis rubric evaluates whether the final report answers the user's analytical request, reports concrete numerical findings, and supports its conclusions with appropriate computations.

\begin{quote}
\textbf{Scoring Rubric for Tabular Data Analysis}

The evaluator is given: (a) the user's analytical request, (b) the final analytical report, defined as the contents of the result editor at the end of the session, and (c) excerpts of the Jupyter execution history that produced the report. The evaluator is instructed to evaluate the final report.

\textbf{Evaluation Criteria}

\textbf{Score = 1.}
There is no report, the report is gibberish, or it is unrelated to the question.

\textbf{Score = 2.}
The report addresses the question only superficially. It contains no specific numerical findings, no statistical justification, unsupported or generic conclusions, or conclusions that contradict the executed code.

\textbf{Score = 3.}
The report answers the question with at least one concrete numerical finding, but lacks rigor. For example, it omits error bars, significance tests, or sample sizes; examines only one slice of the data; or provides conclusions that are only partially supported.

\textbf{Score = 4.}
The report addresses the question with multiple specific numerical findings and uses appropriate aggregations or simple statistics. The conclusions follow from the analysis, with only minor gaps, such as missing caveats about data quality or the absence of a formal hypothesis test where one would have helped.

\textbf{Score = 5.}
The report directly answers the user's question with concrete numbers, uses appropriate statistical testing where relevant, considers alternative explanations or confounders, and presents conclusions with clear caveats. The findings are consistent with the executed code.

The evaluator provides 2--4 sentences of justification and concludes with: \textit{Therefore, the final score is N.}
\end{quote}

\subsection{Visualization Task Rubric}
\label{app:rubric_visualization}

The visualization rubric evaluates the quality of the submitted WildChat-1M analysis, including the hypothesis, code, statistical analysis, visualization, and final interpretation.

\begin{quote}
\textbf{Scoring Rubric for WildChat-1M Analysis}

The evaluator is given: (a) the prompt and constraints, (b) the final code submitted by the participant, and (c) the final output captured at submission. The evaluator is instructed to evaluate the quality of the submitted analysis.

\textbf{Evaluation Criteria}

\textbf{Score = 1.}
No code is submitted, the submission is gibberish, or it is unrelated to the task. This score is also assigned when the submission violates the data-handling constraints in an obvious way, such as pasting raw conversations.

\textbf{Score = 2.}
The code attempts to load WildChat but does not actually compute the hypothesized analysis. There is no plot, only a placeholder plot, no stated hypothesis, or the output is empty or trivially limited to commands such as \texttt{df.head()}.

\textbf{Score = 3.}
The code states a clear hypothesis, runs at least one aggregation or group comparison consistent with the hypothesis, and produces some output. However, statistical testing is missing or weak; the visualization is basic or difficult to interpret; or the conclusions are missing or hand-waved.

\textbf{Score = 4.}
The hypothesis is clear and testable. The code uses appropriate aggregations and at least one statistical test or correlation. The visualization matches the hypothesis and includes proper labels and titles. The output is consistent with the analysis, and the data-handling constraints are respected.

\textbf{Score = 5.}
The hypothesis is well motivated and operationalized. The analysis combines aggregations, an appropriate statistical test with a reported statistic and p-value or effect size, and a polished, well-labeled visualization that supports the conclusion. The code is clean and self-contained, constraints are fully respected, and the conclusion explicitly follows from the results with appropriate caveats.

The evaluator provides 2--4 sentences of justification and concludes with: \textit{Therefore, the final score is N.}
\end{quote}

\section{Likert Correlations }
\label{app:likert_corrs}

In addition to rubric scored task quality, both datasets include subjective Likert ratings collected after the task. These ratings capture participants' perceptions of the outcome and interaction, but they are not interchangeable with our primary quality measure. Table~\ref{tab:likert_corrs} reports pairwise Spearman correlations among the available Likert measures within each dataset.

For CoGym, participants rated the quality of the outcome, the agent, communication with the agent, and overall satisfaction. These measures are moderately to strongly correlated with one another, suggesting that participants' subjective impressions of the interaction are not independent constructs. For example, outcome quality is strongly correlated with both agent rating and communication rating. For the visualization dataset, perceived usefulness, speed, and confidence are also positively correlated, with the strongest relationship between usefulness and speed.

These correlations motivate our decision not to use subjective ratings as the primary quality signal \(Q\). If subjective ratings were used to define outcome quality, later analyses comparing productivity against users' perceived experience would become partially circular. Instead, we use independently rubric scored task performance as the primary quality measure and reserve subjective ratings for comparison analyses. This allows us to test whether users' perceptions track the quality-cost tradeoff captured by our productivity metric.

\begin{table}[t]
\centering
\small
\setlength{\tabcolsep}{4pt}
\begin{tabular}{llrrr}
\toprule
\textbf{Dataset} & \textbf{Rating pair} & \textbf{$\rho$} & \textbf{$p$} & \textbf{$n$} \\
\midrule
CoGym & outcome -- agent & .60 & $<.001$ & 191 \\
CoGym & outcome -- communication & .62 & $<.001$ & 50 \\
CoGym & agent -- communication & .59 & $<.001$ & 50 \\
Website & usefulness -- speed & .67 & $<.001$ & 42 \\
Website & usefulness -- confidence & .56 & $<.001$ & 42 \\
Website & speed -- confidence & .26 & .099 & 42 \\
\bottomrule
\end{tabular}
\caption{Inter-item Spearman correlations among subjective Likert ratings. Subjective ratings are substantially correlated within each study, motivating our decision to use independently rubric-scored task performance as the primary quality signal $Q$ and reserve subjective ratings for comparison analyses.}
\label{tab:likert_corrs}
\end{table}

\begin{figure*}
    \centering
    \includegraphics[width=0.9\linewidth]{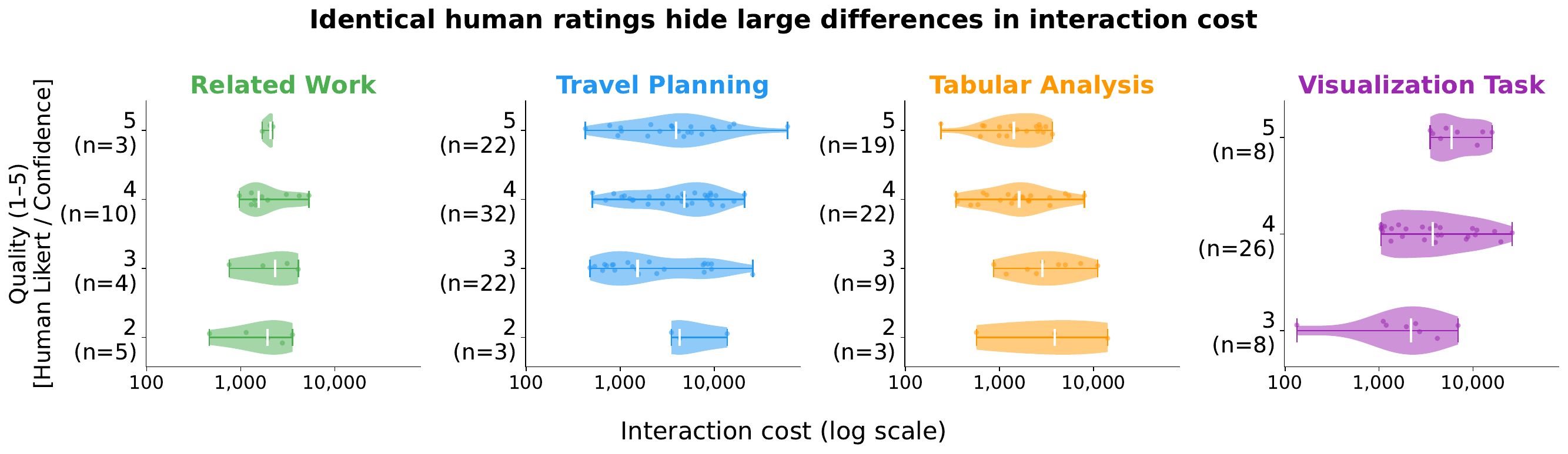}
    \caption{
          Interaction cost within subjective outcome-rating levels. For CoGym, the subjective measure is the participant's outcome quality rating; for the visualization task, it is confidence in the submitted result, since a direct outcome quality rating was not collected. Each point represents one collaboration session, and cost is shown on a log scale. Wide variation within the same rating level shows that subjective outcome ratings can hide large differences in the interaction cost required to produce the final artifact.
    }
    \label{fig:cost_within_quality_likert}
\end{figure*}

\section{Cost Variation Within Human Likert Quality Levels}
\label{app:likert_variation}

Section~\ref{ssec:variation} shows that sessions with identical rubric scored quality can require very different amounts of interaction. Here, we repeat the analysis using the closest available subjective outcome measure in each dataset. For CoGym, this is the participant's Likert rating of outcome quality. For the visualization task, we use confidence in the submitted result as the closest subjective outcome oriented measure.

Figure~\ref{fig:cost_within_quality_likert} plots interaction cost within each human outcome rating level. The same qualitative pattern appears: sessions with the same perceived outcome quality often differ substantially in interaction cost. Even when users assign the same rating to the final output, the amount of interaction required to reach that output can vary widely. This reinforces the motivation for our framework: outcome ratings alone, whether produced by a rubric judge or by the user, do not reveal whether the collaboration was productively achieved or costly to produce. Productivity therefore requires measuring quality and cost jointly rather than treating perceived quality as a complete evaluation signal.

\section{Productivity Rankings Are Robust to Cost and Weighting Choices}
\label{ssec:validate}
% metric validation
We validate our productivity metric: $P_z=z(Q)-z(C)$ along two dimensions. First, we vary the relative weight assigned to agent tokens. Second, we compare productivity rankings under several alternative cost definitions: total tokens, user tokens only, agent tokens only, total turns, and user turns.

\subsection{Varying the Agent-Token Weight}

We define a family of weighted token costs:
\[
C_{\lambda} = \text{user tokens} + \lambda \cdot \text{agent tokens},
\]
where \(\lambda \in \{0.50, 0.75, 1.00, 1.50, 2.00\}\). For each value of \(\lambda\), we recompute productivity as
\[
P_z^{(\lambda)} = z(Q) - z(C_{\lambda}),
\]
with \(Q\) and \(C_{\lambda}\) standardized within task. We then compute pairwise Spearman correlations between the resulting productivity rankings.

\begin{figure}[t]
\centering
\includegraphics[width=\columnwidth]{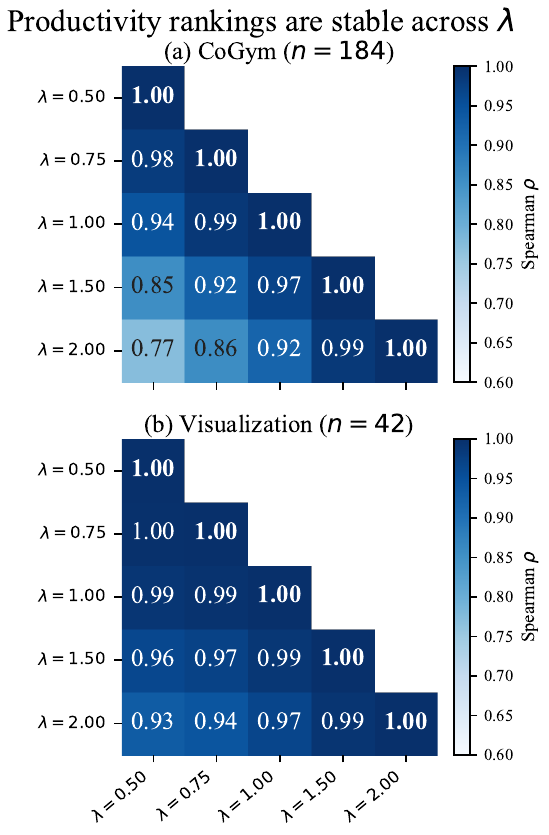}
\caption{Productivity rankings are stable across alternative agent token weights. Each cell reports the Spearman correlation between productivity rankings computed using \(C_{\lambda}=\text{user tokens}+\lambda\cdot\text{agent tokens}\), for \(\lambda \in \{0.50,0.75,1.00,1.50,2.00\}\). The primary setting used in the main text is \(\lambda=.50\). Rankings remain strongly correlated across weights for both CoGym and the visualization dataset.}
\label{fig:lambdas}
\end{figure}

Figure~\ref{fig:lambdas} shows that productivity rankings are stable across agent-token weights. For CoGym, pairwise Spearman correlations range from .77 to 1.00; for the visualization dataset, they range from .93 to 1.00. As expected, the lowest correlations occur when comparing the most different weighting choices, such as \(\lambda=.50\) versus \(\lambda=2.00\). Even in these cases, rankings remain strongly correlated. This suggests that our conclusions are not driven by the specific choice of assigning agent tokens half the weight of user tokens.

\subsection{Alternative Cost Measures}
We also test whether the productivity rankings are robust to broader changes in the cost measure. In addition to \(C_{\text{default}}\), we compute productivity using five alternative cost definitions:
\begin{itemize}[noitemsep]
    \item \textbf{Total tokens}: user tokens + agent tokens.
    \item \textbf{User tokens}: only tokens written by the user.
    \item \textbf{Agent tokens}: only tokens generated by the agent.
    \item \textbf{Total turns}: number of user and agent turns.
    \item \textbf{User turns}: number of user turns.
\end{itemize}

\begin{figure}[t]
\centering
\includegraphics[width=\columnwidth]{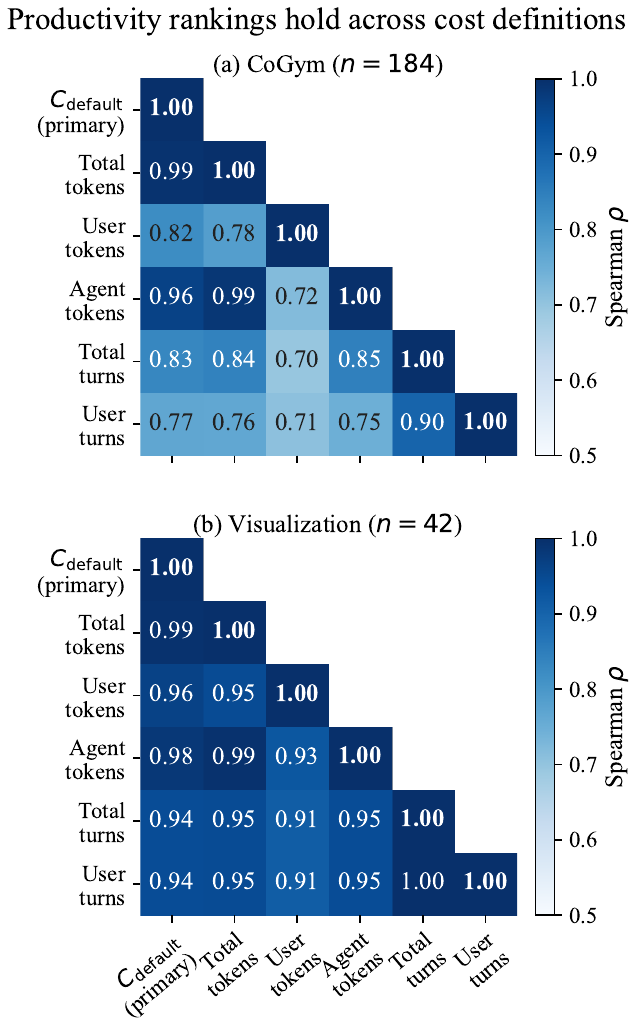}
\caption{Productivity rankings hold across alternative cost definitions. Each cell reports the Spearman correlation between productivity rankings computed with different cost measures. \(C_{\text{default}}\) is the primary weighted token measure used in the main text. Rankings are strongly correlated across token-based and turn-based alternatives, suggesting that the main findings are not an artifact of a single cost operationalization.}
\label{fig:costs}
\end{figure}

Figure~\ref{fig:costs} reports pairwise Spearman correlations between productivity rankings under these cost definitions. The rankings are highly stable. In CoGym, all pairwise correlations are at least .70, and the primary measure is especially close to total tokens (\(\rho=.99\)) and agent tokens (\(\rho=.96\)). In the visualization dataset, all pairwise correlations are at least .91, indicating even stronger agreement across cost definitions.

The somewhat lower correlations for user only measures in CoGym are expected because CoGym agents can take actions in shared task environments and sometimes produce long tool mediated outputs. In these settings, considering only user messages can miss part of the reception burden imposed by the agent. Nevertheless, the overall pattern remains stable that sessions identified as productive under the primary cost definition tend to remain productive under alternative token- and turn-based definitions.

\section{Dialogue Feature Analysis: Full Results}
\label{app:dialogue_full}

Section~\ref{ssec:dialogue} reports the main dialogue patterns associated with productive collaboration. Here, we provide the full feature comparison between the top and bottom productivity quartiles. We compare sessions in the top and bottom quartiles of \(P_z\) across all tasks (\(n=59\) per quartile). For each feature, we compute the mean rate per turn in each quartile, Cohen's \(d\) for the difference in means, and a Mann--Whitney U test. Positive effect sizes indicate features that are more frequent in productive sessions, while negative effect sizes indicate features that are more frequent in unproductive sessions.

The full results in Table~\ref{tab:full_feature_results} support the interpretation in the main text. Productive sessions are characterized by more agent probing and more user acknowledgments or agreements. In contrast, unproductive sessions show more user probing, user assumption reveal, user repair, agent overresponse, and agent overspecification. This pattern suggests that productive collaboration is not defined by the absence of friction. Rather, productivity depends on how grounding work is distributed: productive sessions show agent side clarification followed by user confirmation, while unproductive sessions leave more repair and clarification work to the user.

\begin{table*}[t]
\centering
\small
\setlength{\tabcolsep}{6pt}
\renewcommand{\arraystretch}{1.05}
\begin{tabular}{lrrrr}
\toprule
\textbf{Feature (rate per turn)} & \textbf{Top quartile} & \textbf{Bottom quartile} & \textbf{Cohen's $d$} & \textbf{$p$} \\
\midrule
\multicolumn{5}{l}{\textit{Friction taxonomy} \citep{inan2025better}} \\
\midrule
\texttt{agent probing}            & 0.506 & 0.245 & $+0.73$ & $0.001$\textsuperscript{***} \\
\texttt{probing} (combined)       & 0.423 & 0.344 & $+0.36$ & $0.034$\textsuperscript{*} \\
\texttt{user reinforcement}       & 0.047 & 0.032 & $+0.14$ & $0.783$ \\
\texttt{reinforcement} (combined) & 0.060 & 0.050 & $+0.11$ & $0.650$ \\
\texttt{user reflective pause}    & 0.008 & 0.004 & $+0.09$ & $0.580$ \\
\texttt{agent reinforcement}      & 0.064 & 0.061 & $+0.02$ & $0.425$ \\
\texttt{user overspecification}   & 0.109 & 0.112 & $-0.01$ & $0.162$ \\
\texttt{reflective pause} (combined) & 0.009 & 0.017 & $-0.18$ & $0.155$ \\
\texttt{any friction}             & 0.741 & 0.800 & $-0.31$ & $0.087$ \\
\texttt{overspecification} (combined) & 0.204 & 0.270 & $-0.32$ & $0.020$\textsuperscript{*} \\
\texttt{agent assumption reveal}  & 0.054 & 0.112 & $-0.33$ & $0.115$ \\
\texttt{agent overspecification}  & 0.321 & 0.459 & $-0.38$ & $0.022$\textsuperscript{*} \\
\texttt{agent reflective pause}   & 0.007 & 0.033 & $-0.38$ & $0.187$ \\
\texttt{user probing}             & 0.273 & 0.425 & $-0.52$ & $0.005$\textsuperscript{**} \\
\texttt{user assumption reveal}   & 0.038 & 0.120 & $-0.61$ & $0.000$\textsuperscript{***} \\
\texttt{assumption reveal} (combined) & 0.045 & 0.119 & $-0.66$ & $0.001$\textsuperscript{**} \\
\midrule
\multicolumn{5}{l}{\textit{Grounding act taxonomy} \citep{shaikh2025navigating}} \\
\midrule
\texttt{user acknowledgement}     & 0.165 & 0.049 & $+0.58$ & $0.040$\textsuperscript{*} \\
\texttt{user agreement}           & 0.156 & 0.045 & $+0.58$ & $0.009$\textsuperscript{**} \\
\texttt{user repeat}              & 0.048 & 0.021 & $+0.29$ & $0.922$ \\
\texttt{user disagreement}        & 0.022 & 0.017 & $+0.07$ & $0.349$ \\
\texttt{user topic switch}        & 0.029 & 0.026 & $+0.04$ & $0.309$ \\
\texttt{agent agreement}          & 0.000 & 0.000 & $+0.00$ & $1.000$ \\
\texttt{user clarification}       & 0.054 & 0.067 & $-0.11$ & $0.220$ \\
\texttt{agent clarification}      & 0.025 & 0.042 & $-0.12$ & $0.823$ \\
\texttt{user grounding friction}  & 0.141 & 0.167 & $-0.15$ & $0.257$ \\
\texttt{agent display}            & 0.024 & 0.044 & $-0.22$ & $0.203$ \\
\texttt{agent acknowledgement}    & 0.032 & 0.077 & $-0.35$ & $0.038$\textsuperscript{*} \\
\texttt{user repair}              & 0.039 & 0.080 & $-0.36$ & $0.010$\textsuperscript{*} \\
\texttt{user followup}            & 0.401 & 0.508 & $-0.36$ & $0.072$ \\
\texttt{agent response}           & 0.605 & 0.739 & $-0.41$ & $0.040$\textsuperscript{*} \\
\texttt{agent overresponse}       & 0.469 & 0.620 & $-0.46$ & $0.023$\textsuperscript{*} \\
\bottomrule
\end{tabular}
\caption{Full rate based feature comparison between top and bottom $P_z$ quartiles (n=59 each). Cohen's $d$ is computed on the difference in means; $p$-values from Mann--Whitney $U$ tests. Features marked ``(combined)'' aggregate user and agent turns. Significance: \textsuperscript{*}$p<0.05$, \textsuperscript{**}$p<0.01$, \textsuperscript{***}$p<0.001$.}
\label{tab:full_feature_results}
\end{table*}

\subsection{Per Task Correlations for Agent Probing}
\label{app:per_task_probing}

The main text highlights agent probing as the clearest dialogue feature associated with productive collaboration. Table~\ref{tab:per_task_probing} reports the Spearman correlation between agent probing rate and \(P_z\), both pooled across tasks and separately by task.

The pooled correlation is positive and significant (\(\rho=.19, p=.003\)), but the relationship is task dependent. Agent probing is most strongly associated with productivity in related work writing (\(\rho=.35\)) and travel planning (\(\rho=.24\)), the two tasks where clarification and iterative refinement are often useful. In contrast, the association is near zero in tabular analysis and visualization. This supports the broader claim that productive friction is task relative: agent probing appears most useful when the task benefits from eliciting preferences, constraints, or framing decisions, but less useful when success depends on quickly converging on an executable analysis or visualization.

\begin{table}[h]
\centering
\small
\setlength{\tabcolsep}{6pt}
\begin{tabular}{lrrc}
\toprule
\textbf{Task} & \textbf{$\rho$} & \textbf{$p$} & \textbf{n} \\
\midrule
All tasks           & $+0.19$ & $0.003$\textsuperscript{**} & 233 \\
\midrule
Related work        & $+0.35$ & $0.016$\textsuperscript{*}  & 47  \\
Travel planning           & $+0.24$ & $0.029$\textsuperscript{*}  & 83  \\
Tabular analysis          & $+0.03$ & $0.830$                     & 61  \\
Visualization             & $-0.05$ & $0.731$                     & 42  \\
\bottomrule
\end{tabular}
\caption{Spearman correlations between agent probing rate (\texttt{rate agent probing}) and $P_z$, pooled across all sessions and per-task. The effect is concentrated in tasks where iterative clarification helps (related work, travel planning) and absent in tasks where success depends more on quick convergence (tabular analysis, visualization). \textsuperscript{*}$p<0.05$, \textsuperscript{**}$p<0.01$.}
\label{tab:per_task_probing}
\end{table}

\end{document}